\documentclass[letterpaper]{article}
\usepackage{aaai}
\usepackage{times}
\usepackage{helvet}
\usepackage{courier}

\usepackage{graphicx}
\usepackage{subfigure}
\usepackage{multirow}
\usepackage{amsmath} 
\usepackage{amssymb} 
\usepackage{times}
\usepackage{booktabs}
\usepackage{tabularx}
\usepackage{verbatim}
\begin{document}
%
\title{E2FIF: Push the limit of Binarized Deep Imagery Super-resolution using End-to-end Full-precision Information Flow}
\author{Zhiqiang Lang, Chongxing Song, Lei Zhang, Wei Wei$^{*}$\\
School of Computer Science, Northwestern Polytechnical University, China}
\maketitle
\begin{abstract}
\begin{quote}
Binary neural network (BNN) provides a promising solution to deploy parameter-intensive deep single image super-resolution (SISR) models onto real devices with limited storage and computational resources. To achieve comparable performance with the full-precision counterpart, most existing BNNs for SISR mainly focus on compensating the information loss incurred by binarizing weights and activations in the network through better approximations to the full-precision convolution. In this study, we revisit the difference between BNNs and their full-precision counterparts and argue that the key for good generalization performance of BNNs lies on preserving a complete full-precision information flow as well as an accurate gradient flow passing through each binarized convolution layer. Inspired by this, we propose to introduce a full-precision skip connection or its variant over each binarized convolution layer across the entire network, which can increase the forward expressive capability and the accuracy of back-propagated gradient, thus enhancing the generalization performance. More importantly, such a scheme is applicable to any existing BNN backbones for SISR without introducing any additional computation cost. To testify its efficacy, we evaluate it using four different backbones for SISR on four benchmark datasets and report obviously superior performance over existing BNNs and even some 4-bit competitors.
\end{quote}
\end{abstract}

\section{Introduction}

\begin{figure}[htbp]
  \centering
  \includegraphics[scale=0.8]{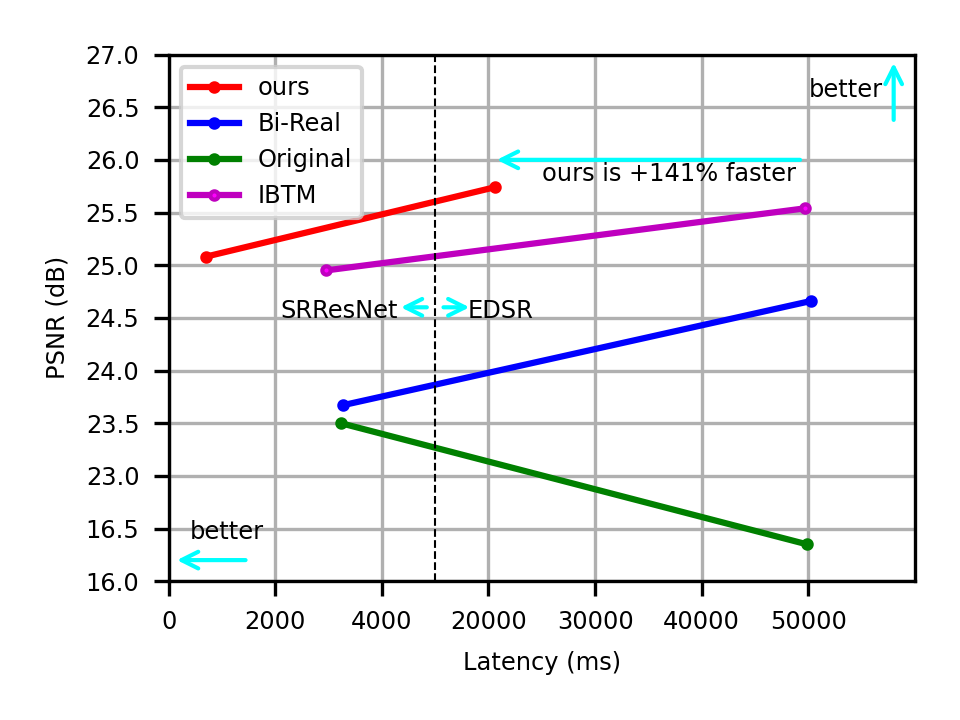}
  \caption{Comparsion with other binarized super-resolution methods(based on $\times$4 scale with SRResNet and EDSR architecture on Urban100). The latencies are tested on Oppo realme GT Master Edition mobile phone equipped with a Qualcomm Snapdragon 870 SoC through Bolt.}
  \label{fig:latency}
\end{figure}

\noindent Deep Convolutional Neural Networks (DCNNs) have achieved impressive performance in many image and video related vision tasks~\cite{he2016deep,lim2017enhanced,wu2021object,qin2021mask,yan2021pixel,wang2021fully}, which however, always demands expensive memory consumption and computational cost. As a result, it is difficult to deploy DCNNs directly on resource-constrained devices. To alleviate this problem, a variety of model compression methods are proposed, among which Binary Neural Networks (BNNs) are well known for their extreme compression and acceleration performance.

Though BNNs have obtained pleasing results on image classification tasks in recent years, the study on BNNs for image super-resolution (SR) is rather limited and can only yield sub-optimal performance. One of the reasons is that the recent studies on SR mainly take advantage of the progress of BNNs for classification, which however, neglects the structure difference between image classification networks and SR networks.

\begin{figure}[t]
  \centering
  \includegraphics[width=0.47\textwidth]{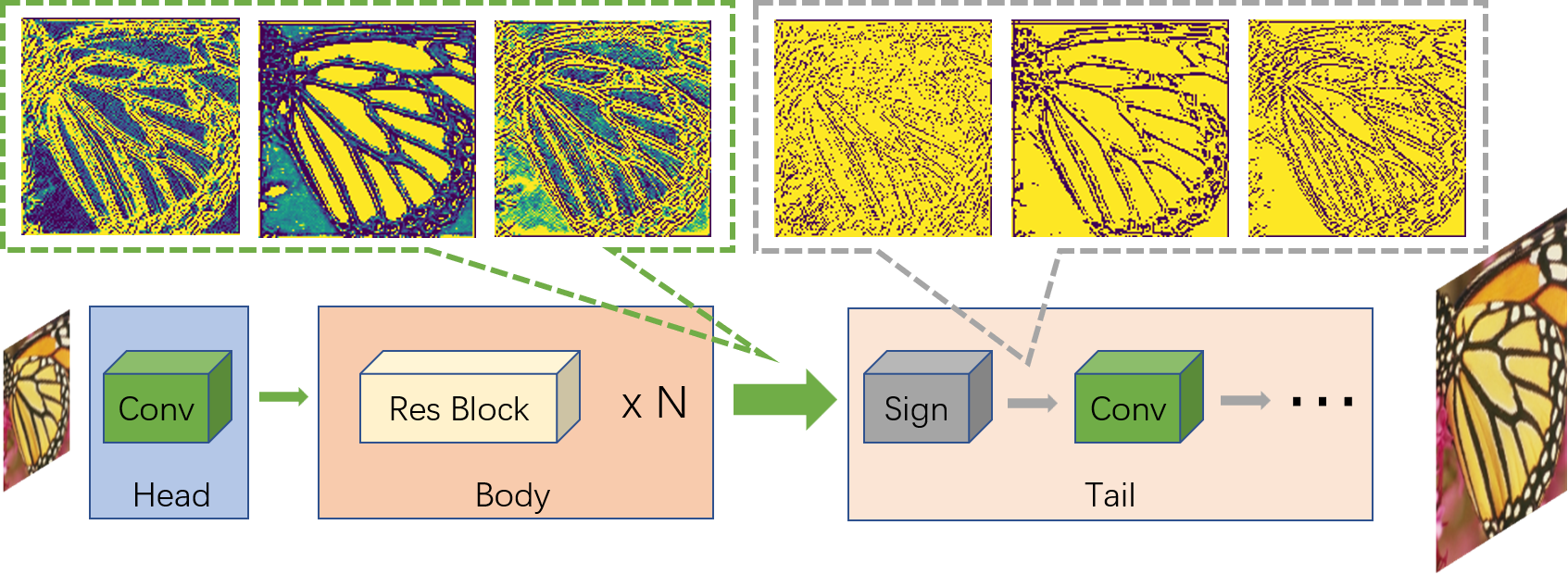}
  \caption{The architecture of a typical binary super-resolution network includes a head module, a body module, and a tail module with two convolutional layers. It can be seen 
  the high-frequency texture will be seriously lost caused by the Sign function within the tail module.}
  \label{fig:story}
\end{figure}

Considering image classification network is composed of numbers of convolutional layer and a fully connected layer, BNNs for image classification mainly focus on compensating the information loss incurred by binarizing weights and activations in the network through better approximations to the full-precision convolution. In contrast, the SR network is more complicated, which 
consists of a head module for initial feature extraction, a body module for detailed feature extraction, and a tail module for upscaling(shown as Figure~\ref{fig:story}). 
Though the networks are different, 
the existing Binary Super-resolution Networks (BSRNs) only pay their attention on the body module which accounts for most of the computational budget, similar as BNNs do for classification. As a result, the tail module is left unnoticed and to be a serious performance bottleneck. Specifically, taking a typical tail module with two convolutional layers shown in Figure~\ref{fig:story} as an example, when binarizing the SR network, the convolutional layer within the head module and the second convolutional layer within the tail module are usually kept from being binarized to guarantee the final performance. Therefore, the tail module starts with a convolutional layer whose input features are binarized. This leads to a severe loss of high-frequency information of the features, which can be seen from the feature maps before and after Sign function in Figure~\ref{fig:story}. As a result, the existing BSRNs can only yield sub-optimal performance. 

The above observation inspires us to rethink BSRNs, and propose new guidelines to construct BSRN suitable for SR structure. Since the bottleneck of existing BSRNs is attributed to the information loss caused by Sign function within the tail module, the efforts on body module only is futile and the integrity of information flow is still destroyed. To tackle with this problem, we propose the first guideline to improve the performance of BSRNs from the perspective of information flow integrity, i.e., \textbf{an end-to-end full-precision information flow (E2FIF) should be able to flow through the entire BSRN}. Following this guideline, we propose two tail modules applicable to BSRNs, including a simple feature repeat shortcut tail module and a lightweight tail module, which can effectively increase the forward expressive capability.  

Besides, considering gradient in back-propagation is also important for BRSNs, we investigate how to devise gradient flow given above mentioned forward-propagate full-precision information flow. For this purpose, we systematically testify different combinations of forward-propagate full-precision information flow and back-propagate gradient flows, from which we obtain the second guideline for BSRNs, i.e., \textbf{the full-precision information flow and the accurate gradient flow can be accessed by each binarized convolutional layer}. Such a conclusion can also provides a deep insight of the Bi-Real Net~\cite{liu2018bi}.

The proposed two guidelines can facilitate us effectively binarize any SR network architectures, from which we build a simple but strong baseline for BSRNs, which is termed as E2FIF and outperforms the state-of-the-art methods with much lower latency(shown as Figure~\ref{fig:latency}).

In summary, the contribution of this study mainly comes from the following aspect.
\begin{itemize}
  \item To the best of our knowledge, we are the first one noticing 
  the tail module destroys the integrity of information flow within BSRNs and becomes the bottleneck, from which we propose to construct BSRNs from the perspective of information flow integrity accordingly. 
  \item We propose two practical guidelines for BSRNs including: 1) an end-to-end full-precision information flow (E2FIF) should be able to flow through the entire BSRN; and 2) the full-precision information flow and the accurate gradient flow can be accessed by
  each binarized convolutional layer. Following these two guidelines, we build a simple but strong baseline for BSRNs termed as E2FIF, which can be adopted to any SR network architectures.
  \item We evaluate E2FIF using four different backbones for SISR
   on four benchmark datasets and report obviously superior performance over existing BNNs and even some 4-bit competitors.
\end{itemize}

\section{Related Works}
\subsection{Gradient Approximation within BNNs}
Non-differentiable Sign function results in the difficulty for training BNNs. As a result, an alternative strategy is to obtain as more as accurate gradients by introducing various artificial prior. For example, \cite{courbariaux2016binarized,hubara2016binarized,rastegari2016xnor} proposed to approximate the gradient of the non-differentiable Sign function by a straight-through estimator (STE). Liu \emph{et al.}~\cite{liu2018bi} further proposed a more accurate approximation function using a piece quadratic function. An artificially designed progressive quantization function is used to reduce the error caused by the estimated gradient. In addition, Zhang \emph{et al.}~\cite{zhang2021embarrassingly} and Xu \emph{et al.}~\cite{xu2021recu} proposed to solve the problem of dead weights caused by tiny gradient by limiting the range of full-precision weights.

\subsection{Representation Capacity of BNNs}
Since the Sign function in BNNs directly quantifies features into $\{-1, +1\}$, it seriously damages the representation capacity respecting its full-precision counterpart. To address this problem, \cite{lin2017towards,zhuang2019structured,liu2019circulant,zhu2019binary}improve the representation capacity by fusing multiple binarized bases to approximate full-precision counterparts. Furthermore, Liu \emph{et al.}~\cite{liu2018bi} proposed to connect with the real activations before a binarized convolutional layer by an identity shortcut. Liu \emph{et al.}~\cite{liu2020reactnet} proposed generalized Sign and PReLU functions with learnable thresholds, enabling explicit learning of the distribution reshape and shift. Similar to \cite{liu2018bi}, we revisit the BNNs from the perspective of information flow, but we focus on the BSRNs, which have different structures from the classification networks. The difference between this study and \cite{liu2018bi} can be seen from the discussion at the end of method.

\subsection{Quantized Super-Resolution Networks}
To apply BNNs to SR tasks, many approaches from different perspectives have been proposed. Li \emph{et al.}~\cite{li2020pams} adopted a truncated parameter to adaptively adjust the upper bound of quantization range for multi-bit SR networks. For BSRNs, Xin \emph{et al.}~\cite{xin2020binarized} proposed a bit-accumulation mechanism to gradually refine features through spatial attention. Jiang \emph{et al.}~\cite{jiang2021training} proposed a new binary training mechanism based on feature distribution for BSRNs, which enables training BSRNs without BN layers. Zhang \emph{et al.}~\cite{zhang2021embarrassingly} proposed a compact uniform prior for the full-precision weights in BSRNs and uses a pixel-level curriculum learning strategy to improve the performance. However, most of these BSRN works draw on the latest advances in BNNs for classification and do not analyze and study the characteristics of BSRNs.

\section{The proposed Method}

In this section, we first revisit the BSRNs from the aspect of information flow integrity and demonstrate the problems of previous methods. We then propose two practical guidelines to construct BSRNs, which can preserve complete full-precision information flow throughout the entire network. 

\subsection{BSRNs Revisited}\label{subsec:rethinking}
Since image classification and image super-resolution are two different tasks, their network structures differ. Specifically, image classification network is composed of numbers of convolutions layer for feature extraction and a fully connected layer for classify the extracted high-level features. In contrast, a typical SR network usually includes a head module contains only a convolutional layer to extract initial features from the low-resolution image, a body module stacks multiple residual blocks for deep feature learning, and a tail module collects the deep features and upscales them to predict the desired high-resolution image.(shown as Figure~\ref{fig:story})

Taking advantages of the recent progress on BNNs for classification, all existing BSRNs pay their attention on the body module, and keep the first and the last convolutional layers within BSRNs from being binarized to guarantee the final performance. However, the tail module is left unnoticed and destroys the integrity of full-precision information flow within BSRNs. Specifically, as shown in Figure~\ref{fig:story}, although the head module and body module keep the full-precision information flow, the Sign function within the tail module binarizes the full-precision feature output by the body module into $\{-1, +1\}$ and leads to a severe loss of  high-frequency information of the features (see the feature maps before and after Sign function in Figure~\ref{fig:story} for details), which makes the tail module be the information bottleneck of BSRNs. As a result, the above observation inspires us to rethink BSRNs from how to effectively preserve the integrity of full-precision information flow and propose two practical guidelines to construct more powerful BSRNs.

\begin{figure}[t]
  \centering
  \subfigure[Original Tail]{
    \includegraphics[scale=0.4]{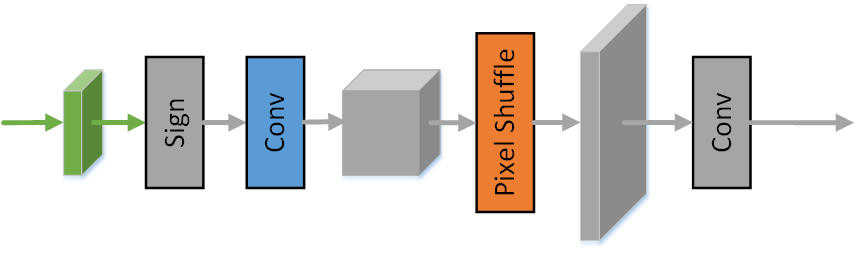}
    \label{fig:originaltail}
  }
  \subfigure[Repeat-Shortcut Tail]{
    \includegraphics[scale=0.4]{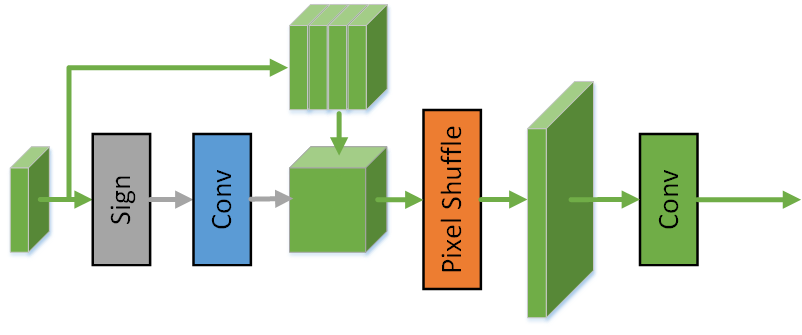}
    \label{fig:repeatshortcuttail}
  }
  \subfigure[Lightweight Tail]{
    \includegraphics[scale=0.4]{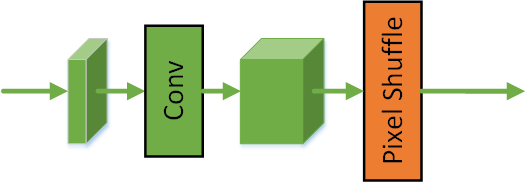}
    \label{fig:lightweighttail}
  }
  \caption{The structure of different tail modules. The gray line represents the discrete information flow with values in a certain range. The green line represents the full-precision information flow. The full-precision information flow is irreversibly destroyed by the Sign function in the Original Tail~\ref{fig:originaltail}. Within the Repeat-Shortcut Tail~\ref{fig:repeatshortcuttail}, the full-precision information flow will be preserved by the repeat shortcut. Within Lightweight tail~\ref{fig:lightweighttail}, the full-precision information flow will be directly utilized to predict the final SR image.}
\end{figure}

\subsection{End-to-End Information Flow guideline for BSRNs}

As discussed above, the information flow bottleneck caused by the Sign function within the tail module limits the performance of BSRNs. As a result, one of the most important issue for constructing more effective BSRNs is to accommodate SR network structure, from which we propose \textbf{\emph{the End-to-End Information Flow guideline}} for BSRNs, i.e., an End-to-end Full-precision Information Flow (E2FIF) should be preserved throughout the entire SR network.    

Following this guideline, we can remodel the existing tail module to preserve the full-precision information flow within the tail module. In this study, we construct two kinds of tail modules suitable for BSRNs accordingly. In the following, we will take the most commonly utilized Original Tail module as an example (shown as Figure~\ref{fig:originaltail}) to clarify how we construct these two kinds of tail modules. 

1) \textbf{\emph{Repeat-Shortcut Tail}}. Since the information flow bottleneck comes from the Sign function, a straightforward way to deal with it is to add a shortcut bypassing the Sign function, similar as that in Bi-Real Net~\cite{liu2018bi}. Repeat-Shortcut tail (Fig.~\ref{fig:repeatshortcuttail}) repeats the input features in the channel dimension and connects them to the output channel-expanded features by the binarized convolutional layer.

2) \textbf{\emph{Lightweight Tail}}.Another way is to drop the Sign function and obtain the lightweight tail which only contains one full-precision convolutional layer, as shown in Fig.~\ref{fig:lightweighttail}. This enables the input full-precision features to be directly used to predict high-resolution images. Though simple, lightweight tail performs even better in experiments compared to Repeat-Shortcut tail.   
\begin{figure}[t]
  \centering
  \subfigure[Orginal Block Forward]{
    \includegraphics[width=0.22\textwidth]{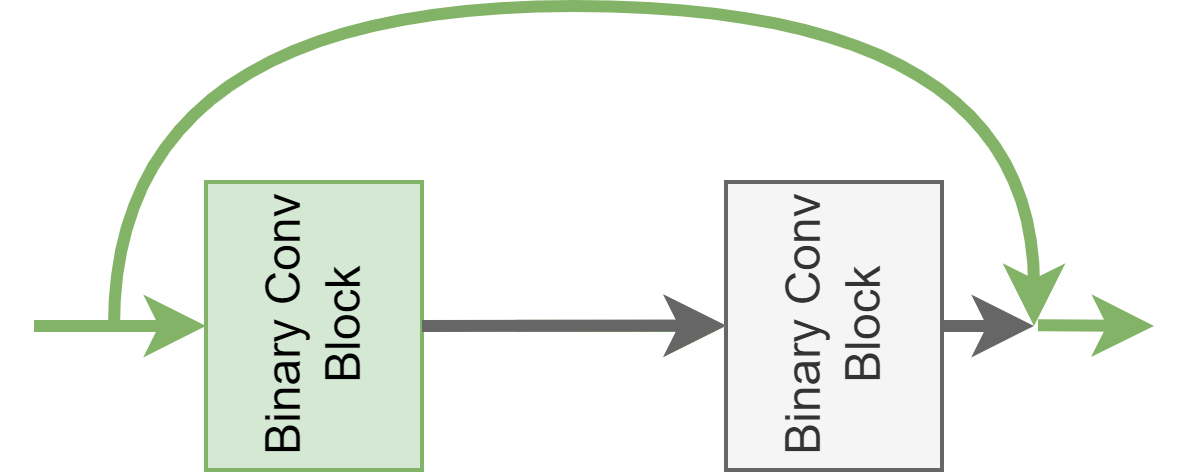}
    \label{subfig:orginal_forward}
  }
  \subfigure[Orginal Block Backward]{
    \includegraphics[width=0.22\textwidth]{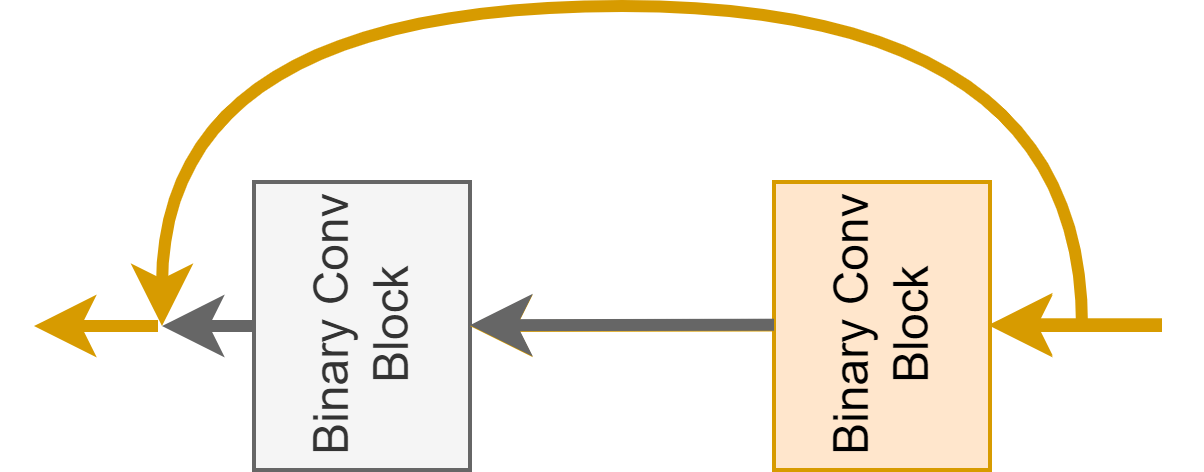}
    \label{subfig:orginal_backward}
  }
  \subfigure[Former Residual Block Forward]{
    \includegraphics[width=0.22\textwidth]{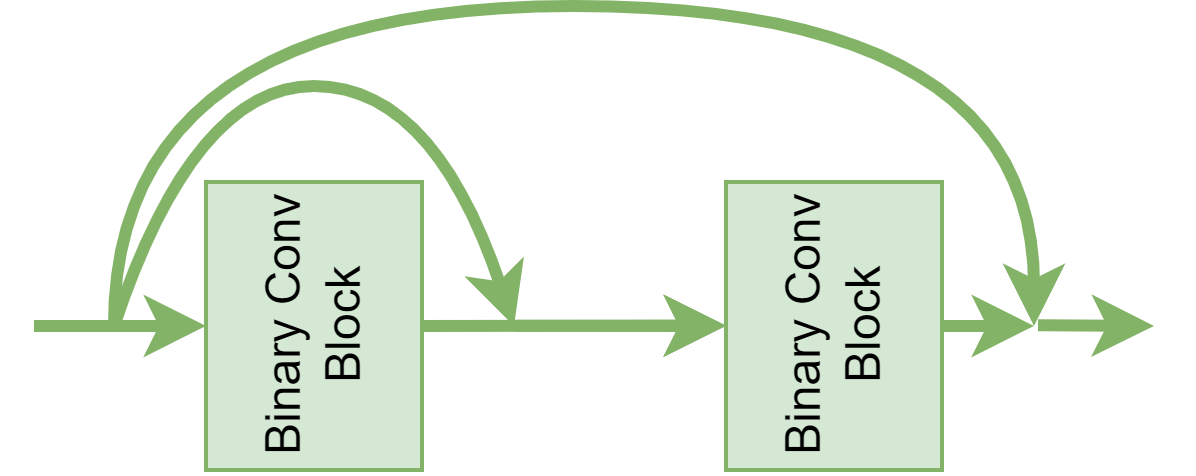}
    \label{subfig:former_forward}
  }
  \subfigure[Former Residual Block Backward]{
    \includegraphics[width=0.22\textwidth]{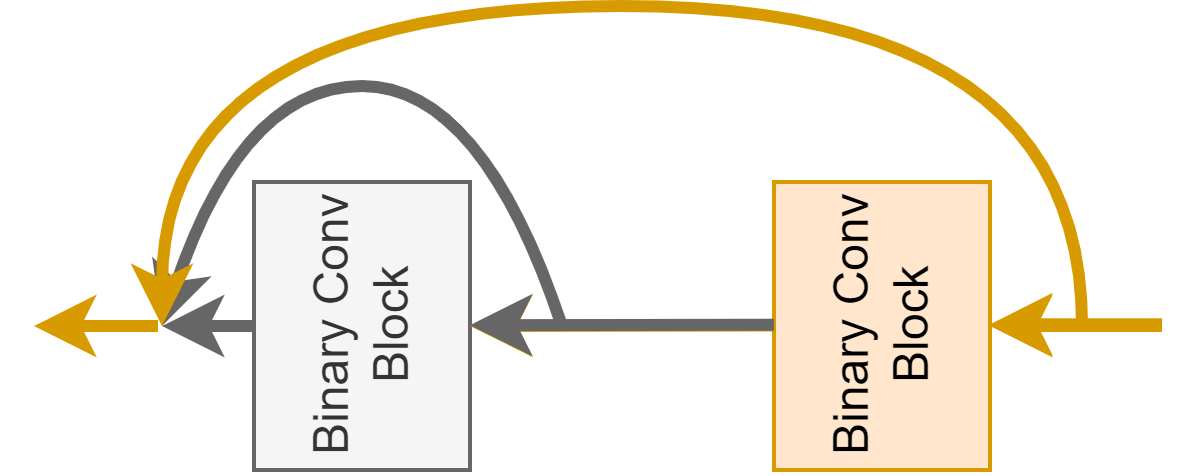}
    \label{subfig:former_backward}
  }
  \subfigure[Later Residual Block Forward]{
    \includegraphics[width=0.22\textwidth]{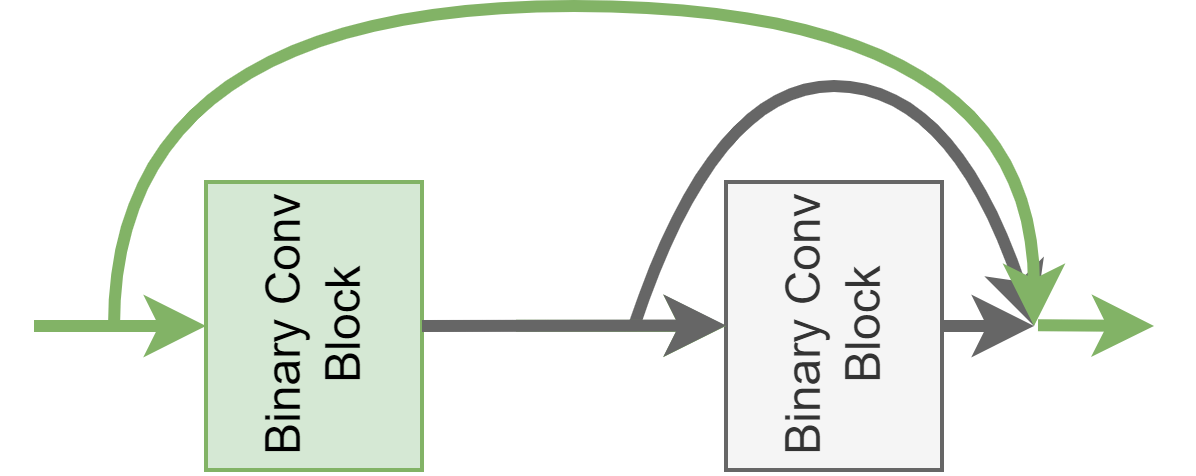}
    \label{subfig:later_forward}
  }
  \subfigure[Later Residual Block Backward]{
    \includegraphics[width=0.22\textwidth]{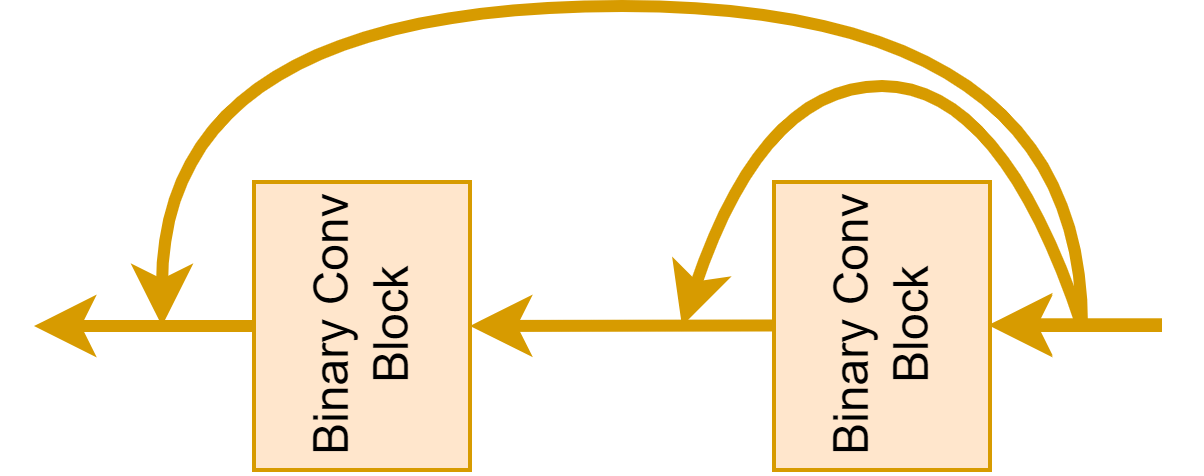}
    \label{subfig:later_backward}
  }
  \subfigure[Bi-Real Block Forward]{
    \includegraphics[width=0.22\textwidth]{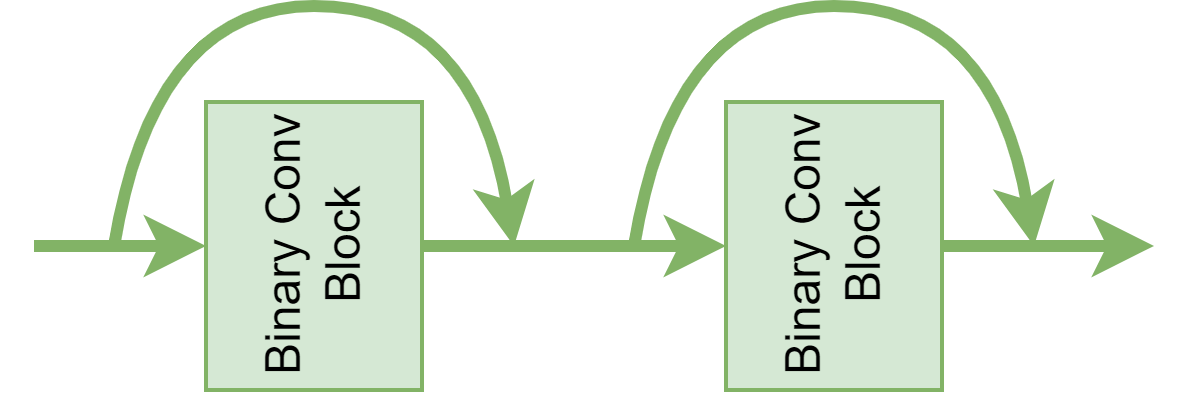}
    \label{subfig:bireal_forward}
  }
  \subfigure[Bi-Real Block Backward]{
    \includegraphics[width=0.22\textwidth]{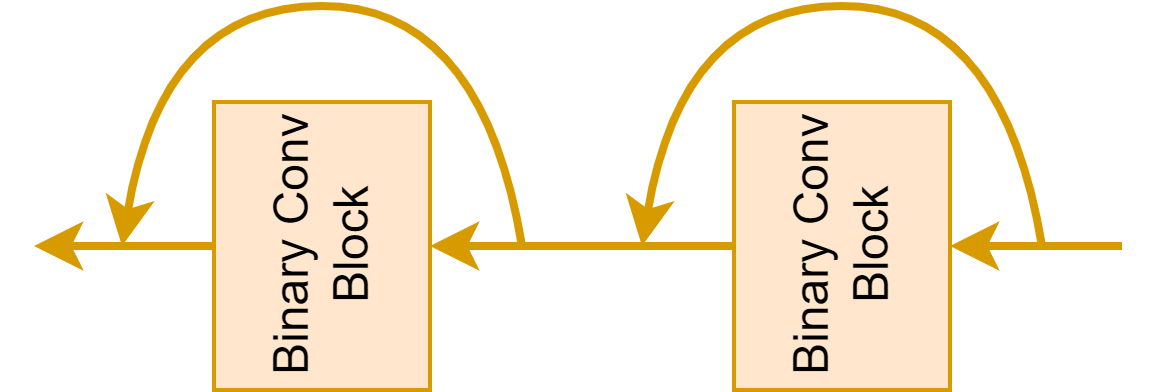}
    \label{subfig:bireal_backward}
  }
  \caption{The forward and backward propagation of four blocks. The four images on the left represent the information flow in the forward propagation of the four modules, in which the green and gray lines represent the full-precision information flow and the discrete information flow, respectively. The four images on the right represent gradient flow in back propagation, w the orange and gray lines represent accurate and inaccurate gradient flow, respectively.}
  \label{fig:block}
\end{figure}

\subsection{Effective Binarized Convolutional Layer guideline for BSRNs}\label{subsec:effectiveconv}

\begin{table*}[t]
  \centering
  \small
  \begin{tabular}{l|c|cc|cc|cc|cc}
  \toprule[1.2pt]
  \multirow{2}{*}{Method} & \multirow{2}{*}{Scale} & \multicolumn{2}{c|}{Set5} & \multicolumn{2}{c|}{Set14} & \multicolumn{2}{c|}{B100} & \multicolumn{2}{c}{Urban100} \\
  & & PSNR & SSSIM & PSNR & SSSIM & PSNR & SSSIM & PSNR & SSSIM \\
  \hline
  \hline
  SRResNet-FullPrecision & $\times$2 & 37.76  & 0.958  & 33.27  & 0.914 & 31.95 & 0.895 & 31.28 & 0.919 \\
  Bicubic & $\times$2 & 33.66  & 0.930  & 30.24  & 0.869 & 29.56 & 0.843 & 26.88 & 0.840 \\
  SRResNet-BNN & $\times$2 & 35.21  & 0.942  & 31.55  & 0.896 & 30.64 & 0.876 & 28.01 & 0.869 \\
  SRResNet-DoReFa & $\times$2 & 36.09  & 0.950  & 32.09  & 0.902 & 31.02 & 0.882 & 28.87 & 0.880 \\
  SRResNet-ABC & $\times$2 & 36.34  & 0.952  & 32.28  & 0.903 & 31.16 & 0.884 & 29.29 & 0.891 \\
  SRResNet-BAM & $\times$2 & 37.21  & 0.956  & 32.74  & 0.910 & 31.60 & 0.891 & 30.20 & 0.906 \\
  SRResNet-E2FIF(ours) & $\times$2 & \textbf{37.50} & \textbf{0.958} & \textbf{32.96} & \textbf{0.911} & \textbf{31.79} & \textbf{0.894} & \textbf{30.73} & \textbf{0.913} \\
  \hline
  \hline
  SRResNet-FullPrecision & $\times$3 & 34.07 & 0.922 & 30.04 & 0.835 & 28.91 & 0.798 & 27.50 & 0.837 \\
  Bicubic & $\times$3 & 30.39 & 0.868 & 27.55 & 0.774 & 27.21 & 0.739 & 24.46 & 0.735 \\
  SRResNet-BNN & $\times$3 & 31.18 & 0.877 & 28.29 & 0.799 & 27.73 & 0.765 & 25.03 & 0.758 \\
  SRResNet-DoReFa & $\times$3 & 32.44 & 0.903 & 28.99 & 0.811 & 28.21 & 0.778 & 25.84 & 0.783 \\
  SRResNet-ABC & $\times$3 & 32.69 & 0.908 & 29.24 & 0.820 & 28.35 & 0.782 & 26.12 & 0.797 \\
  SRResNet-BAM & $\times$3 & 33.33 & 0.915 & 29.63 & 0.827 & 28.61 & 0.790 & 26.69 & 0.816 \\
  SRResNet-E2FIF(ours) & $\times$3 & \textbf{33.65} & \textbf{0.920} & \textbf{29.67} & \textbf{0.830} & \textbf{28.72} & \textbf{0.795} & \textbf{27.01} & \textbf{0.825} \\
  \hline
  \hline
  SRResNet-FullPrecision & $\times$4 & 31.76 & 0.888 & 28.25 & 0.773 & 27.38 & 0.727 & 25.54 & 0.767 \\
  Bicubic & $\times$4 & 28.42 & 0.810 & 26.00 & 0.703 & 25.96 & 0.668 & 23.14 & 0.658 \\
  SRResNet-BNN & $\times$4 & 29.33 & 0.826 & 26.72 & 0.728 & 26.45 & 0.692 & 23.68 & 0.683 \\
  SRResNet-DoReFa & $\times$4 & 30.38 & 0.862 & 27.48 & 0.754 & 26.87 & 0.708 & 24.45 & 0.720 \\
  SRResNet-ABC & $\times$4 & 30.78 & 0.868 & 27.71 & 0.756 & 27.00 & 0.713 & 24.54 & 0.729 \\
  SRResNet-BAM & $\times$4 & 31.24 & 0.878 & \textbf{27.97} & 0.765 & 27.15 & 0.719 & 24.95 & 0.745 \\
  SRResNet-E2FIF(ours) & $\times$4 & \textbf{31.33} & \textbf{0.880} & 27.93 & \textbf{0.766} & \textbf{27.20} & \textbf{0.723} & \textbf{25.08} & \textbf{0.750} \\
  \toprule[1.2pt]
  \end{tabular}
  \caption{The comparison results of different methods on four benchmark datasets at three scales on the SRResNet architecture.}
  \label{table:SRResNet}
\end{table*}

The End-to-End information flow guideline provides an effective way to construct tail modules pertinent for preserving full-precision information flow. In the following, we turn to investigate how to effectively construct body module, still from the perspective of information flow. For this purpose,  we take a commonly utilized structure with two blocks as an example, in which each “Binary Conv Block” (shown in Figure~\ref{fig:block}) follows a “Sign-Conv-Bn” structure. It can be seen there are four kinds of combinations with regards to the full-precision information flow and back-propagate gradient flow, shown as each row in Figure~\ref{fig:block}. The first one is Original Block, in which a shortcut directly over two binarized convolutional layers for both information flow of forward propagation and the gradient flow of backward propagation shown as Figure~\ref{subfig:orginal_forward} and \ref{subfig:orginal_backward}. The second one is Former Residual Block, in which we add extra shortcut over the first binarized convolutional layer into Original Block shown as Figure~\ref{subfig:former_forward} and \ref{subfig:former_backward}.  It is noticeable that Former Residual Block allows the second binarized convolutional layer to additionally receive the full-precision information streams . The third one is Later Residual Block, in which we add extra shortcut over the second binarized convolutional layer into Original Block shown as Figure~\ref{subfig:later_forward} and \ref{subfig:later_backward}. The Later Residual Block allows the first binarized convolutional layer additionally receive an accurate gradient flow. The fourth one is Bi-Real Block, in which each binarized convolutional layer received both the full-precision information and the accurate gradient flow shown as Figure~\ref{subfig:bireal_forward} and \ref{subfig:bireal_backward}.  

Though all these four structures can provide full-precision information flow, the accuracy for them are different, which can be seen in Table~\ref{table:abl_block}. From these results and analysis, we can conclude the accurate gradient flow is beneficial to the full-precision information flow, and propose \textbf{\emph{the Effective Binarized Convolutional Layer guideline}} for BSRNs, i.e., the full-precision information flow and the accurate gradient flow should flow through each binarized convolutional layer as much as possible. We think there are two reasons behind.

Firstly, considering a Res Block with two binarized convolutional layers, with the shortcut over the first binarized convolutional layer, excepting the sign of the input, the magnitude of the input also affects the second binarized convolutional layer, which makes it more sensitive to the input.

Secondly, a part of the accurate gradient can be passed back through the shortcut, which alleviates the backpropagated gradient error caused by the STE, so that the binarized convolutional layers can be better optimized.

\begin{table*}[t]
  \centering
  \small
  \begin{tabular}{l|c|cc|cc|cc|cc}
  \toprule[1.2pt]
  \multirow{2}{*}{Method} & \multirow{2}{*}{Scale} & \multicolumn{2}{c|}{Set5} & \multicolumn{2}{c|}{Set14} & \multicolumn{2}{c|}{B100} & \multicolumn{2}{c}{Urban100} \\
  & & PSNR & SSSIM & PSNR & SSSIM & PSNR & SSSIM & PSNR & SSSIM \\
  \hline
  \hline
  EDSR-FullPrecision & $\times$2 & 38.11 & 0.960 & 33.92 & 0.920 & 32.32 & 0.901 & 32.93 & 0.935 \\
  Bicubic & $\times$2 & 33.66 & 0.930 & 30.24 & 0.869 & 29.56 & 0.843 & 26.88 & 0.840 \\
  EDSR-BNN & $\times$2 & 34.47 & 0.938 & 31.06 & 0.891 & 30.27 & 0.872 & 27.72 & 0.864 \\
  EDSR-BiReal & $\times$2 & 37.13 & 0.956 & 32.73 & 0.909 & 31.54 & 0.891 & 29.94 & 0.903 \\
  EDSR-BNN+ & $\times$2 & 37.49 & 0.958 & 33.00 & 0.912 & 31.76 & 0.893 & 30.49 & 0.911 \\
  EDSR-RTN & $\times$2 & 37.66 & 0.956 & 33.13 & 0.914 & 31.85 & 0.895 & 30.82 & 0.915 \\
  EDSR-BTM & $\times$2 & 37.68 & 0.956 & 33.20 & 0.914 & 31.87 & 0.895 & 30.98 & 0.916 \\
  EDSR-PAMS & $\times$2 & 37.67 & 0.960 & 33.20 & 0.915 & 31.94 & 0.897 & 31.10 & 0.919 \\
  EDSR-IBTM & $\times$2 & 37.80 & 0.960 & \textbf{33.38} & \textbf{0.916} & 32.04 & 0.898 & 31.49 & 0.922 \\
  EDSR-E2FIF(ours) & $\times$2 & \textbf{37.95} & \textbf{0.960} & 33.37 & 0.915 & \textbf{32.13} & \textbf{0.899} & \textbf{31.79} & \textbf{0.924} \\
  \hline
  \hline
  EDSR-FullPrecision & $\times$3 & 34.65 & 0.928 & 30.52 & 0.846 & 29.25 & 0.809 & 28.80 & 0.865 \\
  Bicubic & $\times$3 & 30.39 & 0.868 & 27.55 & 0.774 & 27.21 & 0.739 & 24.46 & 0.735 \\
  EDSR-BNN & $\times$3 & 20.85 & 0.399 & 19.47 & 0.299 & 19.23 & 0.285 & 18.18 & 0.307 \\
  EDSR-BiReal & $\times$3 & 33.17 & 0.914 & 29.53 & 0.826 & 28.53 & 0.790 & 26.46 & 0.801 \\
  EDSR-BNN+ & $\times$3 & 33.56 & 0.919 & 29.73 & 0.831 & 28.68 & 0.794 & 26.80 & 0.820 \\
  EDSR-RTN & $\times$3 & 33.92 & 0.922 & 29.95 & 0.835 & 28.80 & 0.797 & 27.19 & 0.831 \\
  EDSR-BTM & $\times$3 & 33.98 & 0.923 & 30.04 & 0.836 & 28.85 & 0.798 & 27.34 & 0.833 \\
  EDSR-IBTM & $\times$3 & 34.10 & 0.924 & \textbf{30.11} & \textbf{0.838} & 28.93 & 0.801 & 27.49 & 0.839 \\
  EDSR-E2FIF(ours) & $\times$3 & \textbf{34.24} & \textbf{0.925} & 30.06 & 0.837 & \textbf{29.00} & \textbf{0.802} & \textbf{27.84} & \textbf{0.844  } \\
  \hline
  \hline
  EDSR-FullPrecision & $\times$4 & 32.46 & 0.897 & 28.80 & 0.787 & 27.71 & 0.742 & 26.64 & 0.803 \\
  Bicubic & $\times$4 & 28.42 & 0.810 & 26.00 & 0.703 & 25.96 & 0.668 & 23.14 & 0.658 \\
  EDSR-BNN & $\times$4 & 17.53 & 0.188 & 17.51 & 0.160 & 17.15 & 0.151 & 16.35 & 0.163 \\
  EDSR-BiReal & $\times$4 & 30.81 & 0.871 & 27.71 & 0.760 & 27.01 & 0.716 & 24.66 & 0.733 \\
  EDSR-BNN+ & $\times$4 & 31.35 & 0.882 & 28.07 & 0.769 & 27.21 & 0.724 & 25.04 & 0.749 \\
  EDSR-RTN & $\times$4 & 31.49 & 0.884 & 28.14 & 0.771 & 27.27 & 0.726 & 25.20 & 0.756 \\
  EDSR-BTM & $\times$4 & 31.63 & 0.886 & 28.25 & 0.773 & 27.34 & 0.728 & 25.38 & 0.762 \\
  EDSR-PAMS & $\times$4 & 31.59 & 0.885 & 28.20 & 0.773 & 27.32 & 0.728 & 25.32 & 0.762 \\
  EDSR-IBTM & $\times$4 & 31.84 & 0.890 & \textbf{28.33} & \textbf{0.777} & 27.42 & 0.732 & 25.54 & 0.769 \\
  EDSR-E2FIF(ours) & $\times$4 & \textbf{31.91} & \textbf{0.890} & 28.29 & 0.775 & \textbf{27.44} & \textbf{0.731} & \textbf{25.74} & \textbf{0.774} \\
  \toprule[1.2pt]
  \end{tabular}
  \caption{The comparison results of different methods on four benchmark datasets at three scales on  the EDSR architecture.}
  \label{table:EDSR}
\end{table*}

\paragraph{Difference from Bi-Real Net~\cite{liu2018bi}}
The proposed method has similairity with Bi-Real Net in preserving full-precision information flow through shortcuts. But they differ in the following two aspects. 
\textbf{1)} Bi-Real Net only proposed a BNN structure for classification without systematically analyzing the BNNs. In contrast, we systematically analyze the BNNs for SR from the perspective of information flow and propose two guidelines for BSRNs with any complex structures.
\textbf{2)} Bi-Real Net only mentioned that shortcuts can increase the representational capability of the BNNs. But we thought and experimented more deeply on the shortcuts in BNNs, and demonstrated that the accurate gradient flow and the full-precision information flow are equally important for an effective binarized convolutional layer.

\section{Experiments}

In this section, we conduct sufficient comparision experiments and ablation study to demonstrate the effectiveness of the proposed guidelines.

\subsection{Experiments Settings}
\subsubsection{Datasets} We train all models on DIV2K~\cite{agustsson2017ntire} dataset which  contains 800 training images, 100 validation images and 100 testing images. For testing, four benchmark datasets including Set5~\cite{bevilacqua2012low}, Set14~\cite{zeyde2010single}, B100~\cite{martin2001database} and Urban100~\cite{huang2015single} are utilized.

\subsubsection{Evaluation Metrics} Following standard SISR work ~\cite{lim2017enhanced}, PSNR and SSIM are adopted as evaluation metrics. We compare the super-resolution image and the original high-resolution image on the luminance channel Y of the YCbCr color space. The input low-resolution images are generated by the bicubic algorithm.

\subsubsection{Training Settings} 
All experiments are implemented and conducted using the PyTorch framework, on a server platform with 4 V100 GPUs.
All models are trained for 300 epochs from scratch with binary weights and activation. The initial learning rate is set to 2e-4 and halved every 200 epochs. The mini-batch size is set to 16 and the ADAM~\cite{kingma2014adam} optimizer is adapted. 
\subsubsection{Network Architectures and Comparison Methods} To fully demonstrate the effectiveness and generality of the proposed method, we conduct comparisons with state-of-the-art methods on several most commonly utilized network architectures. Specifically, we first conduct comparisons on SRResNet~\cite{ledig2017photo} and EDSR~\cite{lim2017enhanced} architectures previously utilized for BSRNs. The chosen comparison methods include BNN methods such as BNN~\cite{courbariaux2016binarized}, DoReFa Net~\cite{zhou2016dorefa}, ABC Net~\cite{lin2017towards}, Bi-Real Net~\cite{liu2018bi}, BNN+~\cite{darabi2018bnn+} and RTN~\cite{li2020rtn}, toghether with state-of-the-art BSRN methods such as BAM~\cite{xin2020binarized} and IBTM~\cite{jiang2021training}, multi-bit quantization SR network PAMS~\cite{li2020pams}.

In addition to the SRResNet and EDSR architectures, we also binarize two advanced SR architectures including RCAN~\cite{zhang2018image} and RDN~\cite{zhang2018residual}. Since none of the previous methods have attempt to binarize these two architectures, we mainly compare the proposed method with our baseline method including Bi-Real Net~\cite{liu2018bi} and the recent state-of-the-art method IBTM~\cite{jiang2021training}.

\subsection{Comparison of Quantitative Results}
\subsubsection{Results on SRResNet Architecture~\cite{ledig2017photo}}
The results of all comparison methods on four benchmark datasets are shown in Table~\ref{table:SRResNet}. The proposed E2FIF obtains the best performance on all scales and datasets, except PSNR of our method is only 0.04 dB slightly lower than that from BAM on Set14 at 4x scale. 
But it is noticeable more extra computations are utilized in BAM due to the bit accumulation mechanism for activations,
Compared with ABC which approximates the full-precision convolution through multiple binarized convolutions, the proposed E2FIF improves the PSNR over 1 dB at 2x scale. All these results demonstrate the effectiveness of the proposed E2FIF.

\subsubsection{Results on EDSR Architecture~\cite{lim2017enhanced}}
EDSR has a similar structure as SRResNet, but with more Residual Blocks and channels. 
As can be seen from Table \ref{table:EDSR}, the proposed E2FIF achieves best performance on three datasets. 
Though the performance of the proposed E2FIF is only slightly lower than IBTM on Set14, it has a clear advantages over IBTM as well as other methods on the other three datasets including the most difficult dataset Urban100.
Compared with Bi-Real Net which can be considered as the baseline of our method, the proposed E2FIF achieves more than 1 dB improvement at all settings with nearly the same amount of computations.
More importantly, the proposed E2FIF also has obvious advantages compared with the 4-bit quantized SR network PAMS, which also can demonstrated the importance of the proposed guidelines for BSRNs.
\subsubsection{Results on RCAN Architecture~\cite{zhang2018residual} and RDN Architecture~\cite{zhang2018residual}}
The proposed E2FIF also performs best on these two architectures compared to other methods, demonstrating the compatibility of the proposed guidelines with complex structures. Due to the length limitation of this paper, we provide detailed experiments results and analyses in the supplementary material.

\subsection{Model Analysis}
\begin{figure*}[t]
  \begin{center}
  \subfigure[Bicubic]{\includegraphics[width=0.16\textwidth,angle=0]{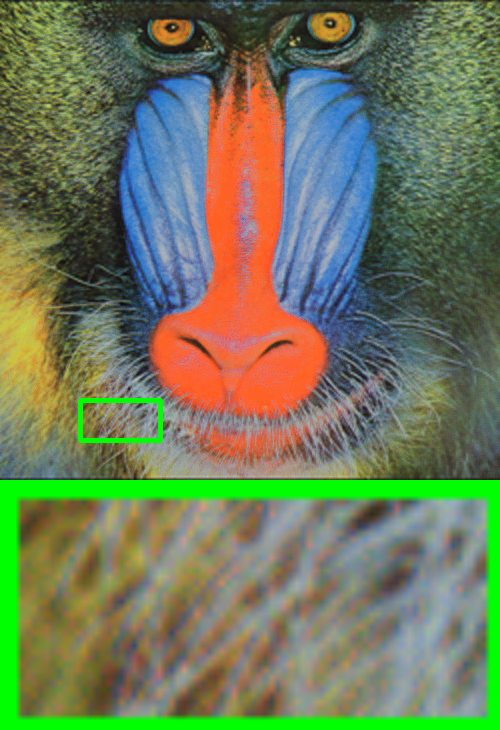}}
  \subfigure[Original Tail]{\includegraphics[width=0.16\textwidth,angle=0]{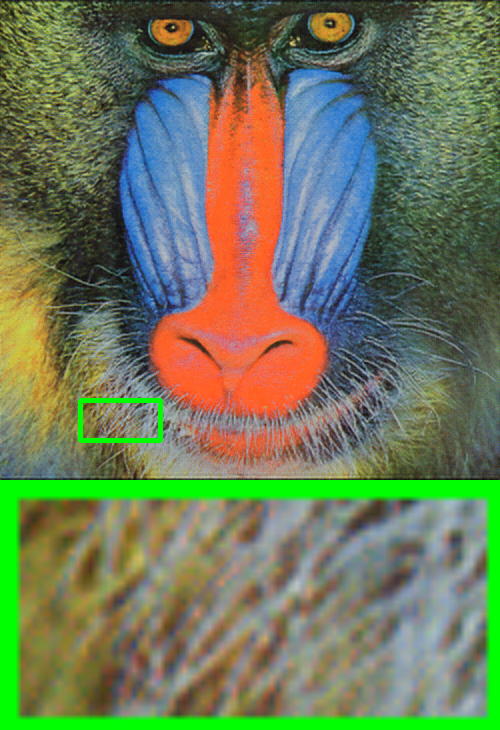}}
  \subfigure[Repeat-Shortcut Tail]{\includegraphics[width=0.16\textwidth,angle=0]{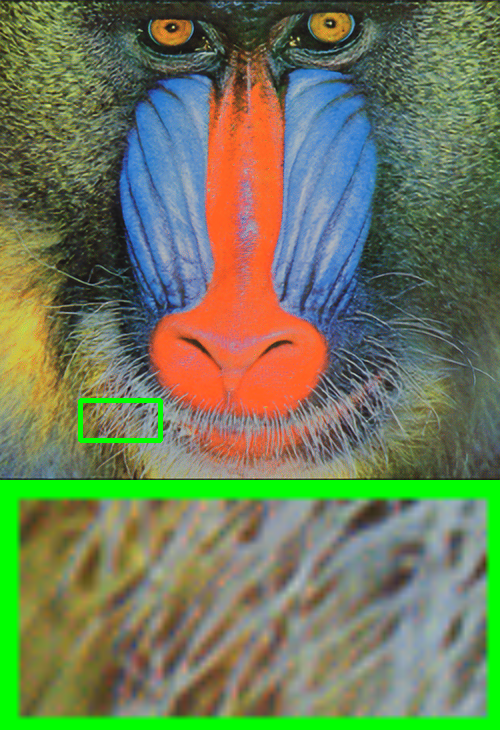}}
  \subfigure[Lightweight Tail]{\includegraphics[width=0.16\textwidth,angle=0]{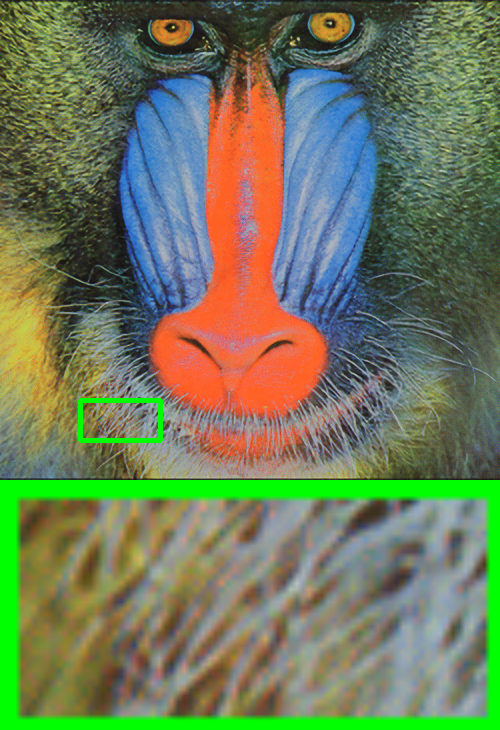}}
  \subfigure[Full Precision]{\includegraphics[width=0.16\textwidth,angle=0]{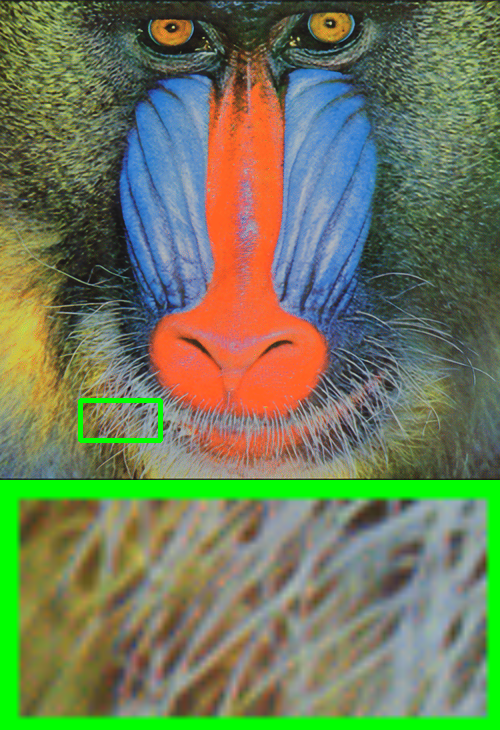}}
  \subfigure[Ground Truth]{\includegraphics[width=0.16\textwidth,angle=0]{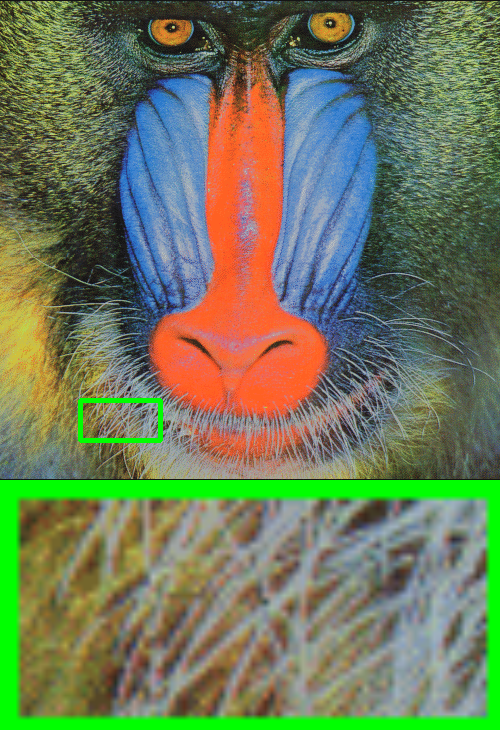}}
  \end{center}
  \caption{Visual results from Set14 at $\times$2 super resolution with the SRResNet~\cite{ledig2017photo} architecture.}
  \label{fig:x2}
\end{figure*}

\begin{figure*}[t]
  \begin{center}
  \subfigure[Bicubic]{\includegraphics[width=0.138\textwidth,angle=0]{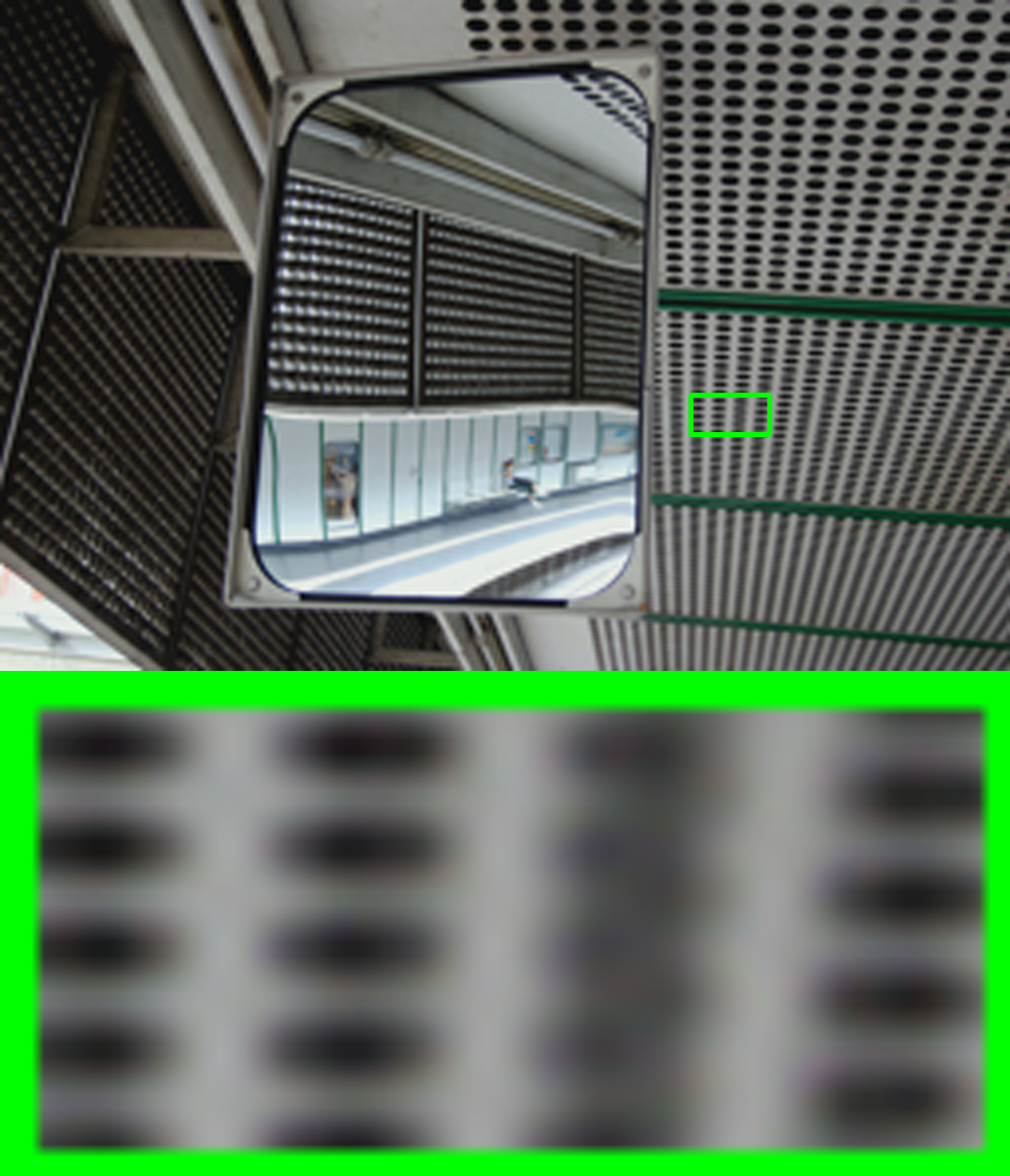}}
  \subfigure[Bi-Real]{\includegraphics[width=0.138\textwidth,angle=0]{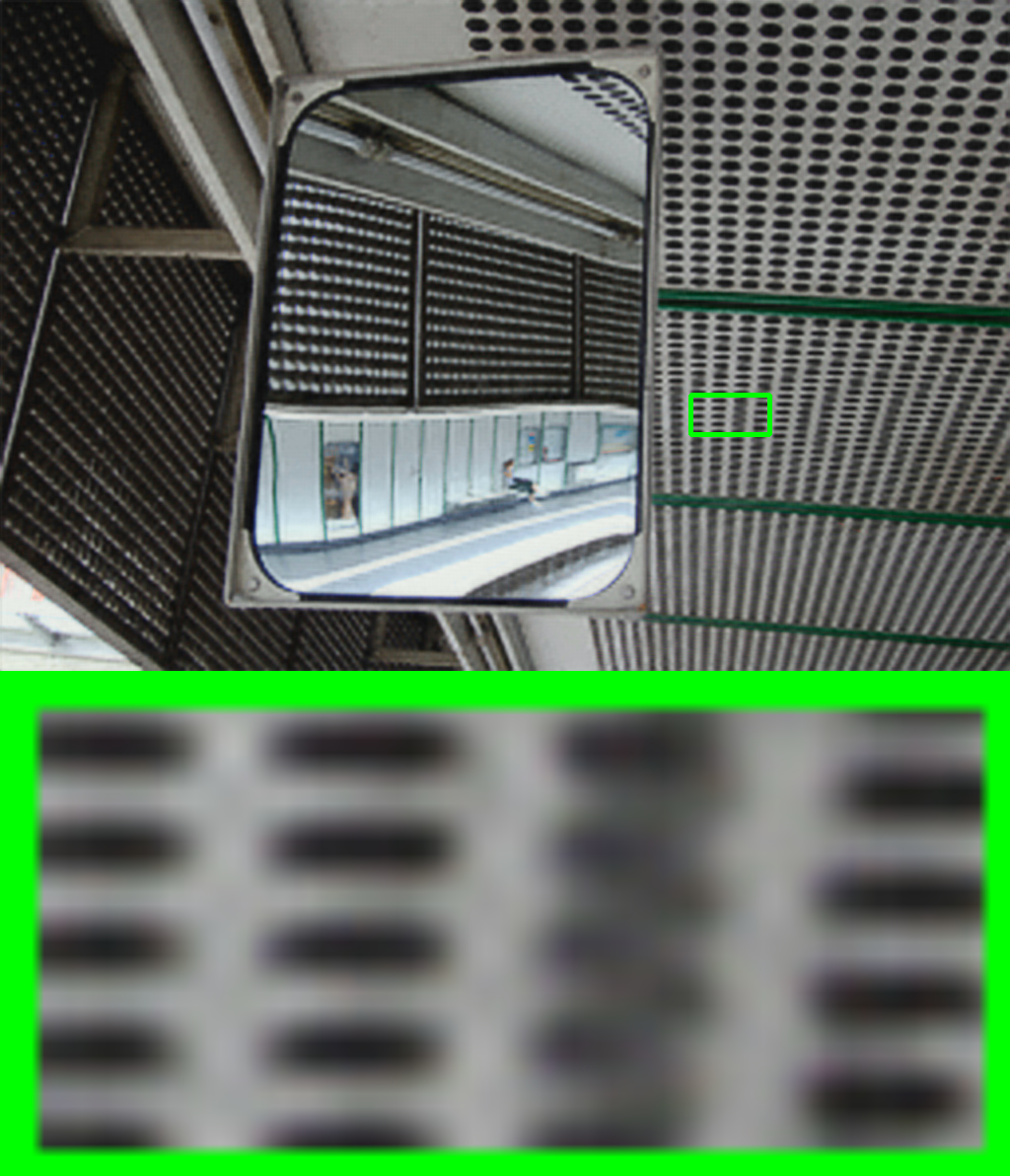}}
  \subfigure[Variant1]{\includegraphics[width=0.138\textwidth,angle=0]{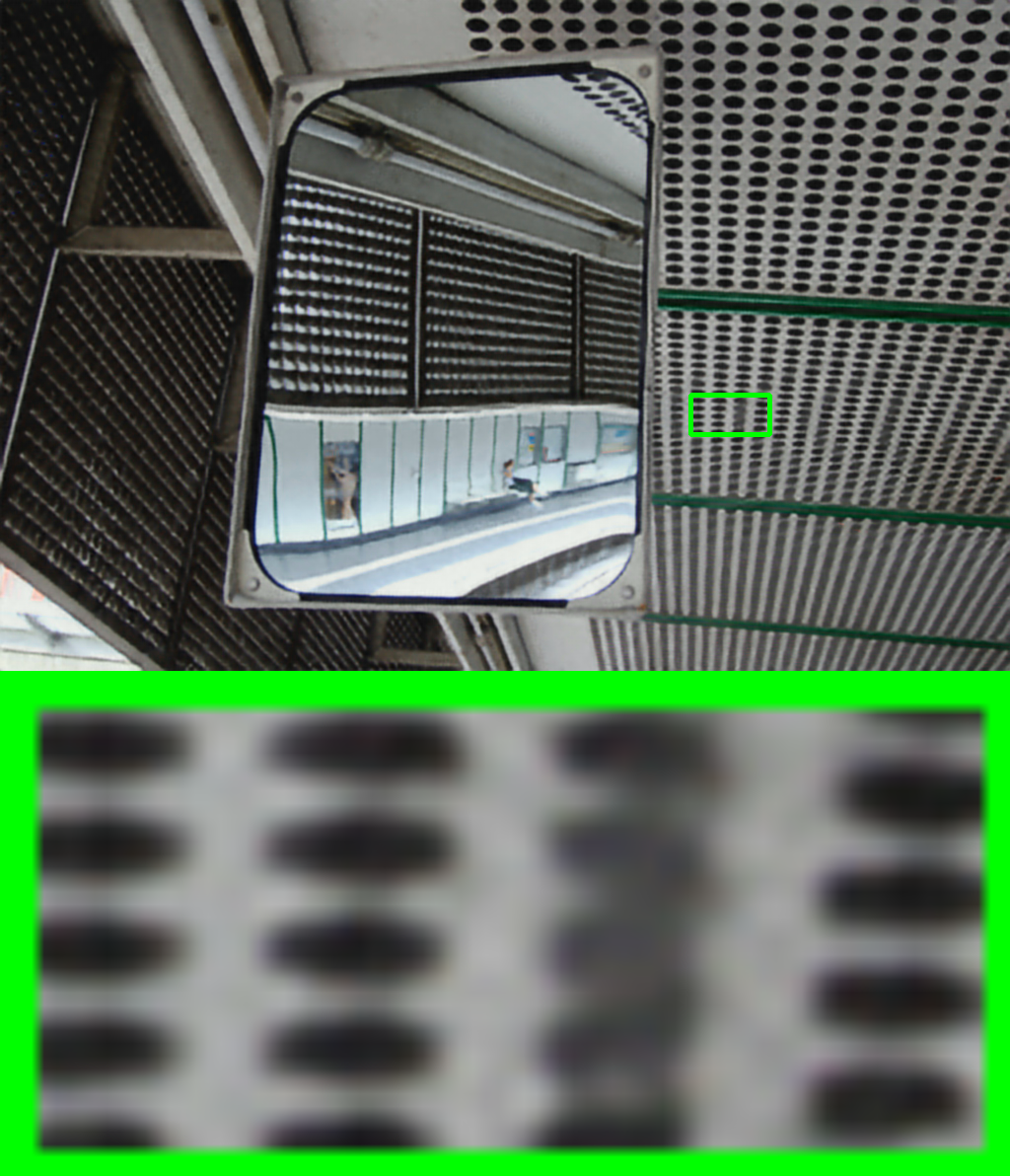}}
  \subfigure[Variant2]{\includegraphics[width=0.138\textwidth,angle=0]{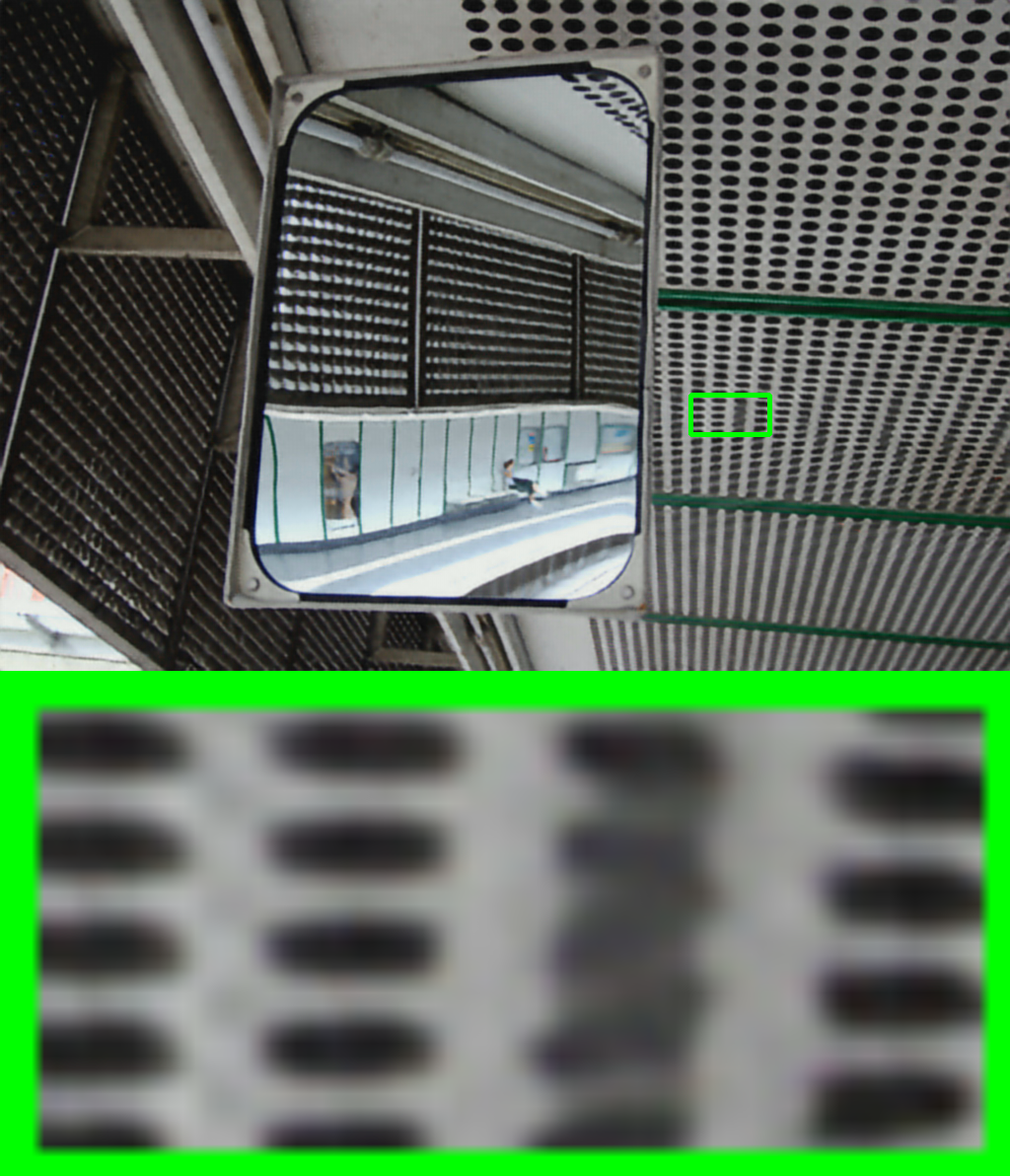}}
  \subfigure[E2FIF]{\includegraphics[width=0.138\textwidth,angle=0]{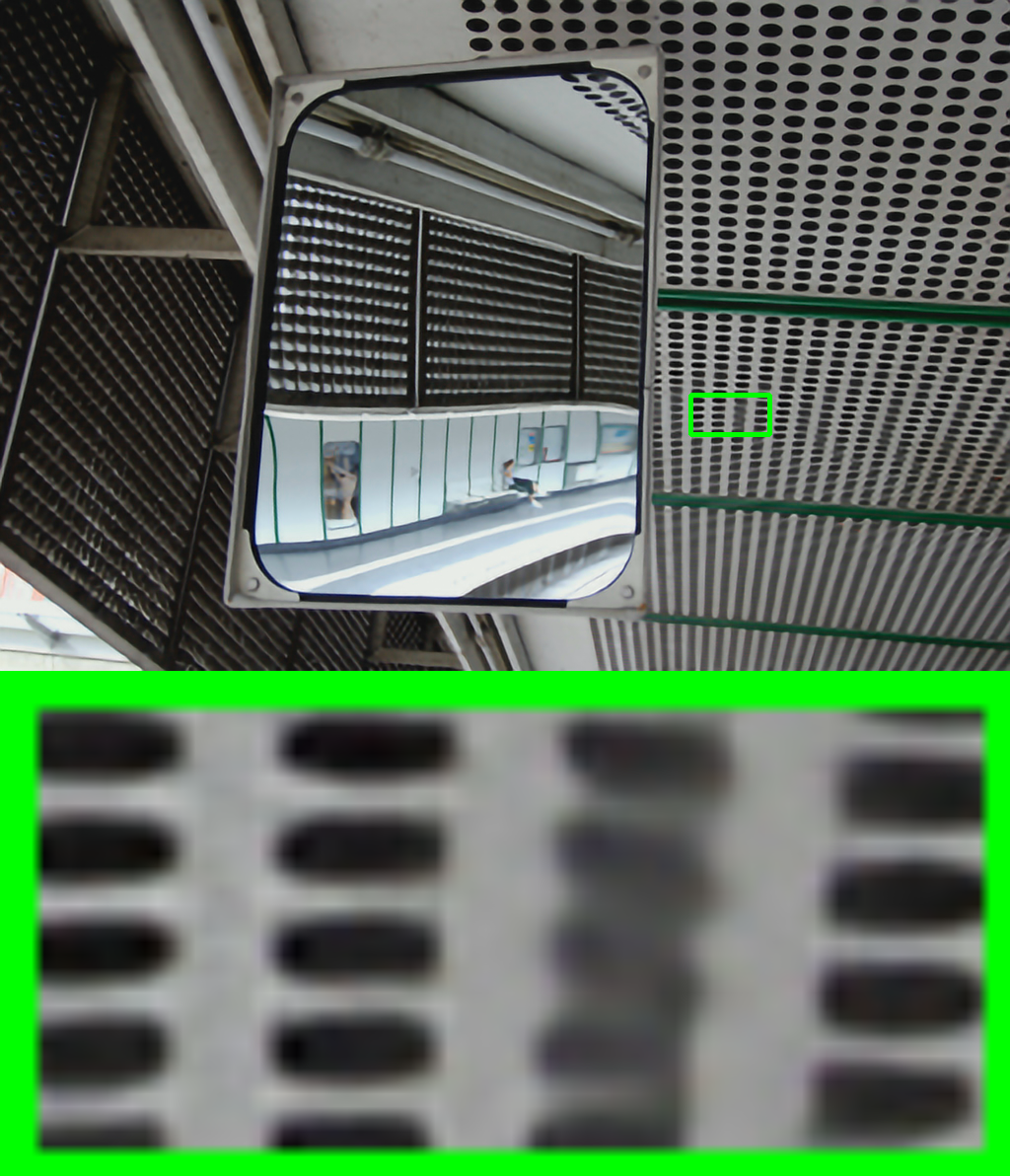}}
  \subfigure[Full precision]{\includegraphics[width=0.138\textwidth,angle=0]{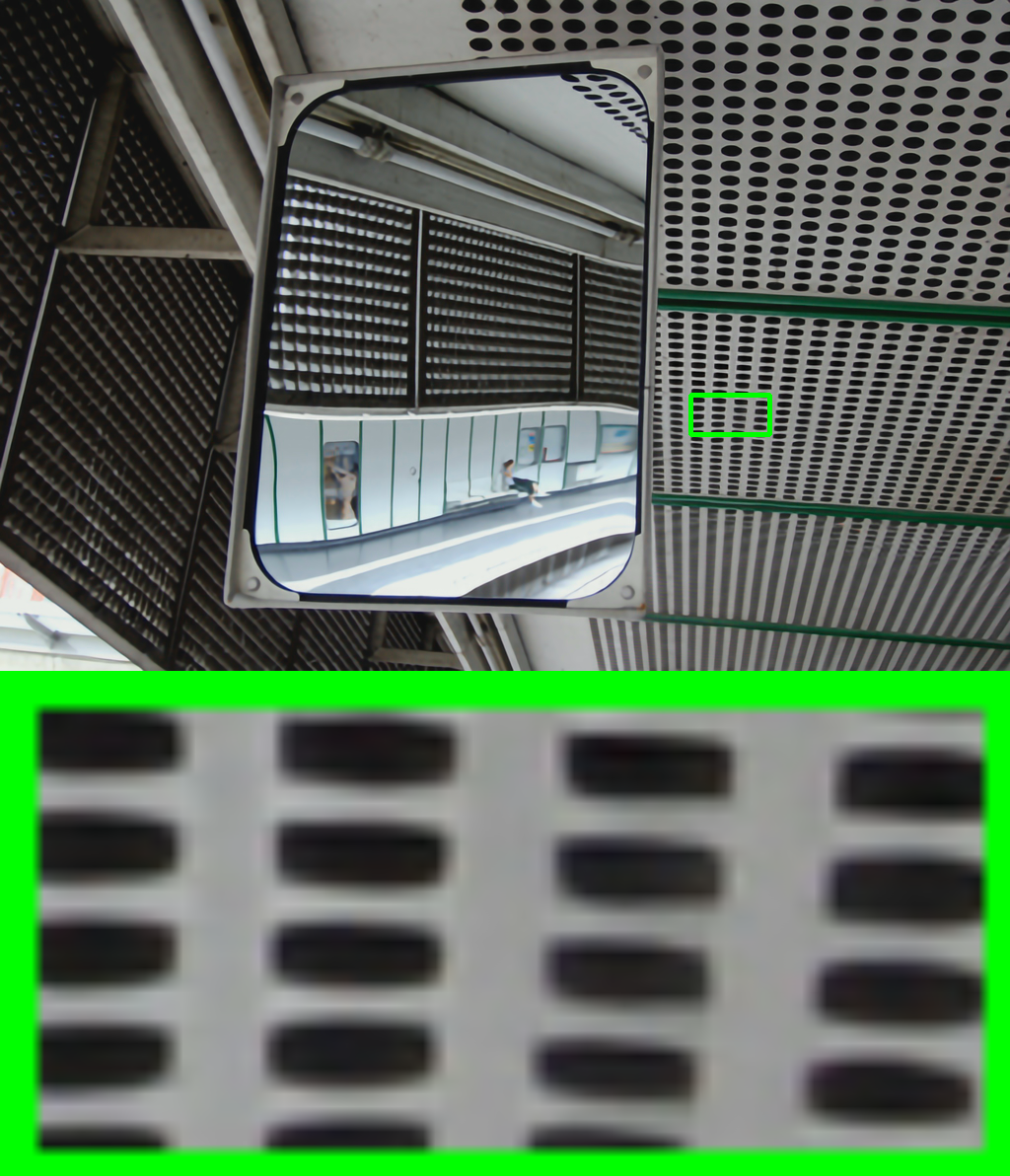}}
  \subfigure[Ground Truth]{\includegraphics[width=0.138\textwidth,angle=0]{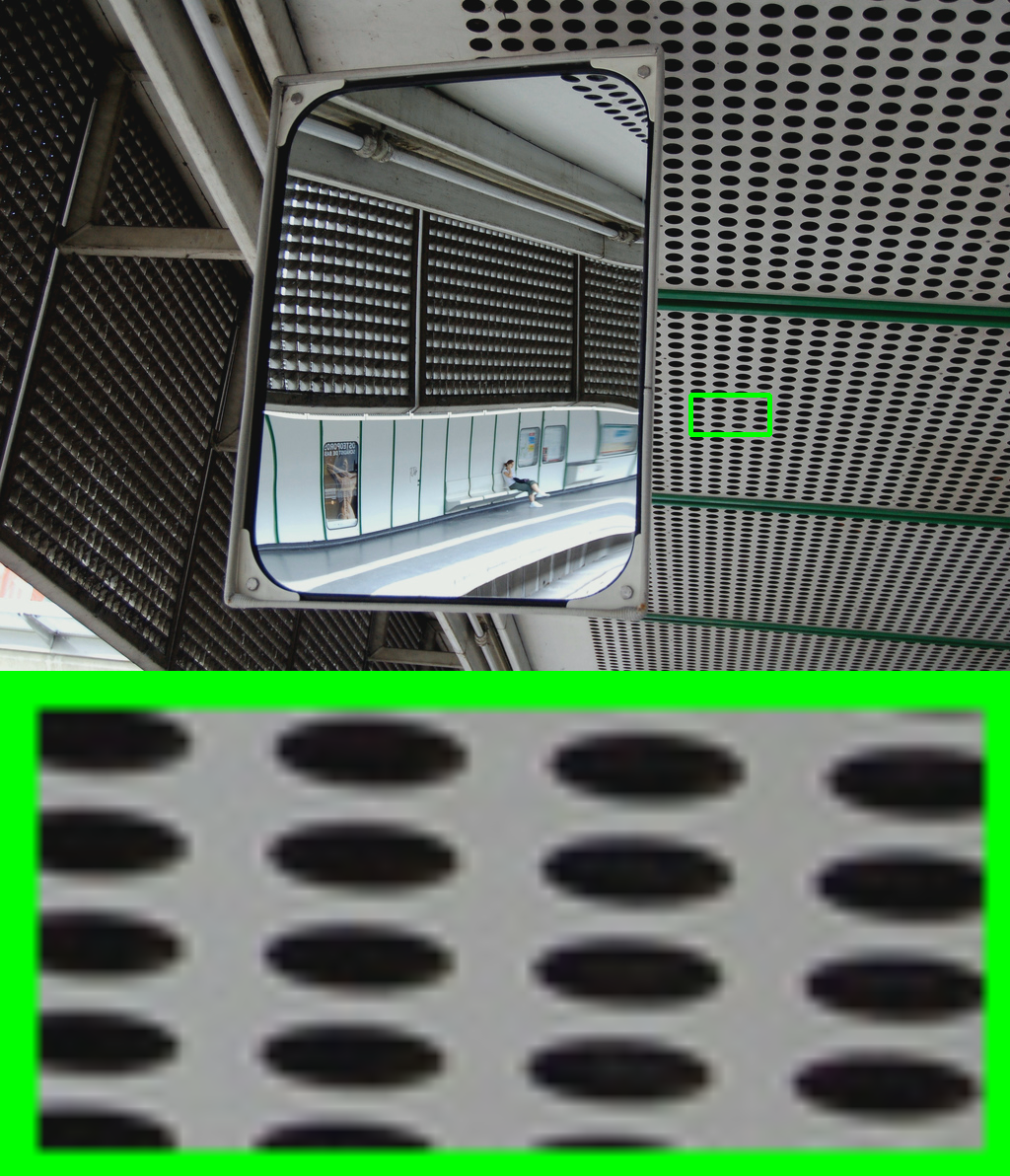}}
  \end{center}
  \caption{Visual results from Urban100 at $\times$4 super resolution with the RCAN~\cite{zhang2018image} architecture. The meanings of Variant1 and Variant2 are introduced in the caption of Table 1 in supplementary material.}
  \label{fig:x4}
\end{figure*}

\subsubsection{Ablation study of different tail module}\label{subsubsec:tail}
To verify the effectiveness of the proposed Repeat-Shortcut and Lightweight Tails, we compare them with the Original Tail module on SRResNet architecture, as shown in Table~\ref{table:tail_abl}. It can be seen that the performance of the network is effectively improved by simplely adding a repeat shortcut to the Original Tail. Furthermore, the proposed Lightweight Tail directly removes the first binarized convolutional layer, which futher reduces the information flow loss and improves the performance as well as reduces the computation cost. These effectively demonstrate the applicability of the proposed Lightweight Tail and the importance of the E2FIF guideline for BSRNs.
\begin{table}[t]
  \centering
  \small
  \begin{tabular}{c|cc}
  \toprule[1.2pt]
  Tail & PSNR & SSIM \\
  \hline
  \hline
  Original Tail & 23.67 & 0.673 \\
  Repeat-Shortcut Tail & 24.62 & 0.733 \\
  Lightweight Tail & \textbf{25.08} & \textbf{0.750} \\
  \hline
  \hline
  Full Precision & 25.54 & 0.767 \\
  \toprule[1.2pt]
  \end{tabular}
  \caption{The results of binarized SRResNet with different tail modules on Urban100 at $\times$4 super resolution.}
  \label{table:tail_abl}
\end{table}

\subsubsection{Analysis of different Block}\label{subsubsec:block}
\begin{table}[t]
  \centering
  \small
  \begin{tabular}{c|cccc}
  \toprule[1.2pt]
  Metrics & Original & Former & Later & Bi-Real \\
  \hline
  \hline
  PSNR & 24.86 & 24.92 & 24.95 & \textbf{25.08} \\
  SSIM & 0.741 & 0.744 & 0.745 & \textbf{0.750} \\
  \toprule[1.2pt]
  \end{tabular}
  \caption{The results of binarized SRResNet with different blocks on Urban100 at $\times$4 super resolution.}
  \label{table:abl_block}
\end{table}

The results of binarized SRResNet with different block is shown as Table~\ref{table:abl_block}. As can be seen, the former and later shortcut added into the Original Tail improved the performance, which demonstrate the importance of the full-precision information flow and the accurate gradient flow. More importantly, the performance of Bi-Real Net whose each convolutional layer received both full-precision information flow and accurate gradient flow is further improved, demonstrating the compatibility of the effective binarized convolutional layer guideline for BSRNs.

\subsubsection{Analysis of cutoff at different position}
\begin{table}[t]
  \setlength\tabcolsep{3pt}
  \centering
  \small
  \begin{tabular}{c|cccccc|c}
  \toprule[1.2pt]
  \multirow{2}{*}{Metrics} & \multicolumn{6}{c|}{Body} & \multirow{2}{*}{Tail} \\
  & 0 & 8 & 16 & 24 & 30 & 31 & \\
  \hline
  \hline
  PSNR & 24.91 & 25.00 & 25.03 & 24.99 & 24.80 & 24.66 & 23.67 \\
  SSIM & 0.743 & 0.747 & 0.748 & 0.745 & 0.740 & 0.735 & 0.673 \\
  \toprule[1.2pt]
  \end{tabular}
  \caption{The results of binarized SRResNet with cutoff at different position on Urban100 at $\times$4 super resolution.}
  \label{table:abl_cutoff}
\end{table}

The effect of cutoffs at different position of the binarized SRResNet is shown as Table~\ref{table:abl_cutoff}. As can be seen, more close to the beginning and end of the network, the performance will the  hurts more by cutoff. In addition, the cutoff of the tail has the largest impact on the performance. This is consistent with our conjectures that the truncated information flow will be gradually restored by the following layers. However, a cutoff at the beginning will cause the initial information flow to be damaged, and a cutoff at the end will cause the information flow to be too lated to be restored.
\subsection{Comparison of Deployment Efficiency}
As shown in Figure~\ref{fig:latency}, we compare the inference latency and performance of the proposed E2FIF as well as other methods on Oppo realme GT Master Edition mobile phone equipped with a Qualcomm Snapdragon 870 SoC through Bolt, an efficient binary network inference framework. The proposed method achieved the highest PSNR with half latency due to the lightweight tail module when testing on Urban100 at x4 scale on both SRResNet and EDSR architecture. This is mainly because the designed lightweight tail discards the convolution operation performed on high-resolution feature maps compared to the original tail, which reduces the computation and enhances the speed remarkably.  
\subsection{Comparison of Qualitative Results}
We visualize the reconstruction results on SRResNet architecture in Figure~\ref{fig:x2} for comparison. As can be seen, the reconstruction results of Repeat-Shortcut Tail and Lightweight Tail much better than that from the Original Tail, and even has no obvious difference with the visual results from its Full-precision counterpart. In addition, we also visualize the reconstruction results at $\times$4 super resolution on RCAN architecture of different variants. As can be seen from Figure~\ref{fig:x4}, the edges of the super-resolution images reconstructed by the proposed E2FIF is more sharper, and more closer to the result from its full-precision counterpart.

\section{Conclusion}

In this work, we systematically analyze the BSRNs from an information flow perspective and proposed two guidelines for BSRNs. Firstly, preserving the integrity of the end-to-end full-precision information flow is necessary for the BSRNs. Secondly, the accurate gradient flow and the full-precision information flow are equally important for an effective binarized convolutional layer. The proposed E2FIF based on the guidelines achieves state-of-the-art performance with adding little computational cost. More importantly, we can effectively binarize any complex SR network with the proposed guidelines.


\bibliographystyle{aaai}
\bibliography{formatting-instructions-latex}

\end{document}